\pdfoutput=1

\documentclass[11pt]{article}

\usepackage{EMNLP2022}

\usepackage{times}
\usepackage{latexsym}

\usepackage[T1]{fontenc}

\usepackage[utf8]{inputenc}

\usepackage{microtype}

\usepackage{graphicx} 

\usepackage{inconsolata}

\title{Few-shot Query-Focused Summarization with Prefix-Merging}

\author{Ruifeng Yuan \\
The Hong Kong Polytechnic University \\
  \texttt{csryuan@comp.polyu.edu.hk} \And
  Zili Wang \\
  Xidian University \\
  \texttt{ziliwang.do@gmail.com} \AND
  Ziqiang Cao \footnotemark[1]\\
  Institute of Artificial Intelligence, \\Soochow University, China \\
  \texttt{zqcao@suda.edu.cn} \And
  Wenjie Li \\
The Hong Kong Polytechnic University \\
  \texttt{cswjli@comp.polyu.edu.hk} \\}

\begin{document}
\maketitle
\begin{abstract}
Query-focused summarization has been considered as an important extension for text summarization. It aims to generate a concise highlight for a given query. Different from text summarization, query-focused summarization has long been plagued by the problem of lacking high-quality large-scale datasets. In this paper, we investigate the idea that whether we can integrate and transfer the knowledge of text summarization and question answering to assist the few-shot learning in query-focused summarization. Here, we propose prefix-merging, a prefix-based pretraining strategy for few-shot learning in query-focused summarization. Drawn inspiration from prefix-tuning, we are allowed to integrate the task knowledge from text summarization and question answering into a properly designed prefix and apply the merged prefix to query-focused summarization. With only a small amount of trainable parameters, prefix-merging outperforms fine-tuning on query-focused summarization. We further discuss the influence of different prefix designs and propose a visualized explanation for how prefix-merging works. 
\end{abstract}
\footnotetext[1]{Corresponding author}
\section{Introduction}

Text summarization aims to compress the source document(s) into a shorter version that contains its important information. As a classic sub-topic for text summarization, query-focused summarization meets that situation that only a specific aspect of information is needed to be summarized. In other words, it aims to generate a summary based on the source content related to a given query. Hence, this task requires not only to locate relevant content in a passage as question answering (QA) but also to summarize and generate a highlight as text summarization. Although text summarization has been widely studied in recent years, there are fewer attempts on exploring query-focused summarization \cite{deng2020multi,su2020caire,xu2020coarse,su2021improve} after the age of neural model. One main reason is the lack of generalized large-scale datasets. Compared with the easily accessible nature reference summaries such as titles or headlines in text summarization, it is hard to collect large-scale data for query-focused summarization. Meanwhile, human-written reference summaries have always been costly.

The rapidly developed few-shot learning techniques provides potential cues to alleviate the problem of lacking large-scale dataset for query-focused summarization, and knowledge transferring is one of them. In fact, when facing unseen tasks, it is natural for human beings to integrate and transfer the knowledge of known tasks to relevant new tasks. Inspired by this, we innovatively propose to decouple the query-focused summarization to two basic tasks, i.e. text summarization and question answering, and transfer the knowledge from these two tasks to query-focused summarization. However, in parameter-based knowledge learning, previous work are usually one-to-one (pre-train then fine-tune \cite{yosinski2014transferable}) or one-to-many (domain/task adaption \cite{houlsby2019parameter,lin2020exploring}), and seldom of them focus on many-to-one (integrate basic tasks to a complex one). In this case, the previous methods may not work well in this task.

In this paper, we propose a pre-trained strategy, prefix-merging, for few-shot learning in query-focused summarization. In recent prompt-based language models, the prompt/prefix is considered as containing the knowledge of the given task, which provides us an explicit way to control the task-specific knowledge previously dispersed in the language model (LM). For example, prefix-tuning \cite{li2021prefix} achieved a similar result with fine-tuning by training only the task-specific prefix, a sequence of continuous vectors that prepend to the input. Following the framework proposed by prefix-tuning, prefix-merging aims to integrate the task knowledge from text summarization and question answering into a properly designed prefix and apply the merged prefix to the more complex task, query-focused summarization.

Generally, there are two straightforward ideas for merging knowledge from multiple tasks into a prefix: concatenate the separated prefix for different tasks as a whole or adopt a shared prefix for all the tasks. Considering there exist both similarities and differences across the tasks, a more flexible prefix design composed of both task-specific part and shared part is used in further investigation. Moreover, we propose a self-adaptive prefix merging that allows the basic tasks themselves to decide the prefix design. Drawn the inspiration from \cite{xu2021raise}, we adopt Fisher Information to calculate the importance scores of the prefix embeddings (basic units for the prefix) for each basic task. For one task, only the prefix embeddings with top scores are activated in the following training. Hence, different tasks can adapt to different parts of prefix automatically. After finishing training the merged prefix, it is transferred to a downstream task for few-shot learning. In the experiment, we explore prefix merging in the context of query-focused summarization, taking PubMedQA \cite{jin2019PubMedQA} and DUC \cite{dang2006duc} as the evaluation dataset.

Prefix-merging provides a potential solution for the few-shot learning in complex tasks that can be integrated by the basic tasks. Benefited by the universality of the prompt-based approach, prefix-merging is not limited by the model architecture and can be used in both autoregressive LM and encoder-decoder based LM. We believe this shows a possible direction to the application of prompt-based approaches. Our contribution can be summarized as follow:
\begin{itemize}
	\item We provide a new solution for few-shot query-focused summarization by decoupling it to two basic tasks with large-scale training data, text summarization and question answering.
	\item We propose prefix-merging that integrates the task-specific knowledge from basic tasks to assist the learning of a more complex task, which provides a new solution to many-to-one parameter-transfer learning.
	\item We further expand the application of prompt-based approaches by applying the prefix to multi-task situation, exploring the interaction between different task knowledge through prefix.

\end{itemize}

\section{Related Work}
\subsection{Query-focused Summarization}

Query-focused summarization aims to generate a concise highlight from the source document(s) according to a specific topic or query, which is considered as a more complex extension of text summarization. Early works \cite{lin2010putting,shen2011learning} focus on extracting  query-related sentences as summaries, while further works \cite{wang2016sentence,li2014query} improve it by rewriting the extracted sentences with sentence compression. \cite{nema2017diversity,hasselqvist2017query} propose neural-abstractive models with an additional query attention mechanism to generate the summaries with respect to the given query. \cite{deng2020multi} consider the relation among the query and source sentences as a multi-hop inference process and generate the summaries by integrating information from different inference steps. Meanwhile, researchers also utilized QA models to find the possible query-related evidence in query-focused summarization. \cite{xu2020coarse,xu2020generating} adopts QA models for sentence-level or paragraph-level answer evidence ranking. \cite{su2021improve} incorporate answer relevance scores generated by QA model as explicit fine-grained query relevance to a transformer-based abstractive summarization model. Therefore, we believe the text summarization and QA are the foundation for query-focused summarization and choose them as the auxiliary tasks in this work. 

\subsection{Prompt-based Approaches}
Prompting originally refers to adding instructions and several examples to a task input and generating the output from the LM. A fundamental idea for prompt-based approaches is that let the tasks adapt to the LM. Some researchers tend to utilize the idea to improve the performance of the model by making the form of the task closer to the LM. A series of works \cite{petroni2019language,jiang2020can,shin2020autoprompt} explore the prompt engineering and prompt ensemble in natural language understanding tasks. For instance, instead of manually designing prompt, AutoPrompt \cite{shin2020autoprompt} automatically search for a sequence of discrete words as prompt to extract knowledge from pre-trained LMs. Other works choose to optimize the prompt in a continuous space. \cite{qin2021learning,liu2021gpt} adopt hand-designed prompt as initialization and add learnable perturbation on the prompt. Other researchers choose to find a parameter-efficient adaption from LM to a specific task. GPT-3 \cite{brown2020language} adopts manually designed task-specific prompts to adapt the LM for different generation tasks. Prefix-tuning proposes “prefix tuning” for language generation task: learning a sequence of continuous prefixes that are inserted to every transformer layer. \cite{lester2021power} provides a simplified version of “prefix tuning” with fewer parameters and more robust prompt initialization on the SuperGLUE tasks. \cite{zhao2022domain} has recently proposed a prefix-based model that utilize domain words to achieve zero-shot domain adaption on dialogue summarization.
In this work, following the framework of prefix-tuning, we aim to integrate basic tasks to a more complex one by merging the task knowledge through the prefix.

\section{Method}

\begin{figure*}[ht]
	\centering
	\includegraphics[scale=0.85]{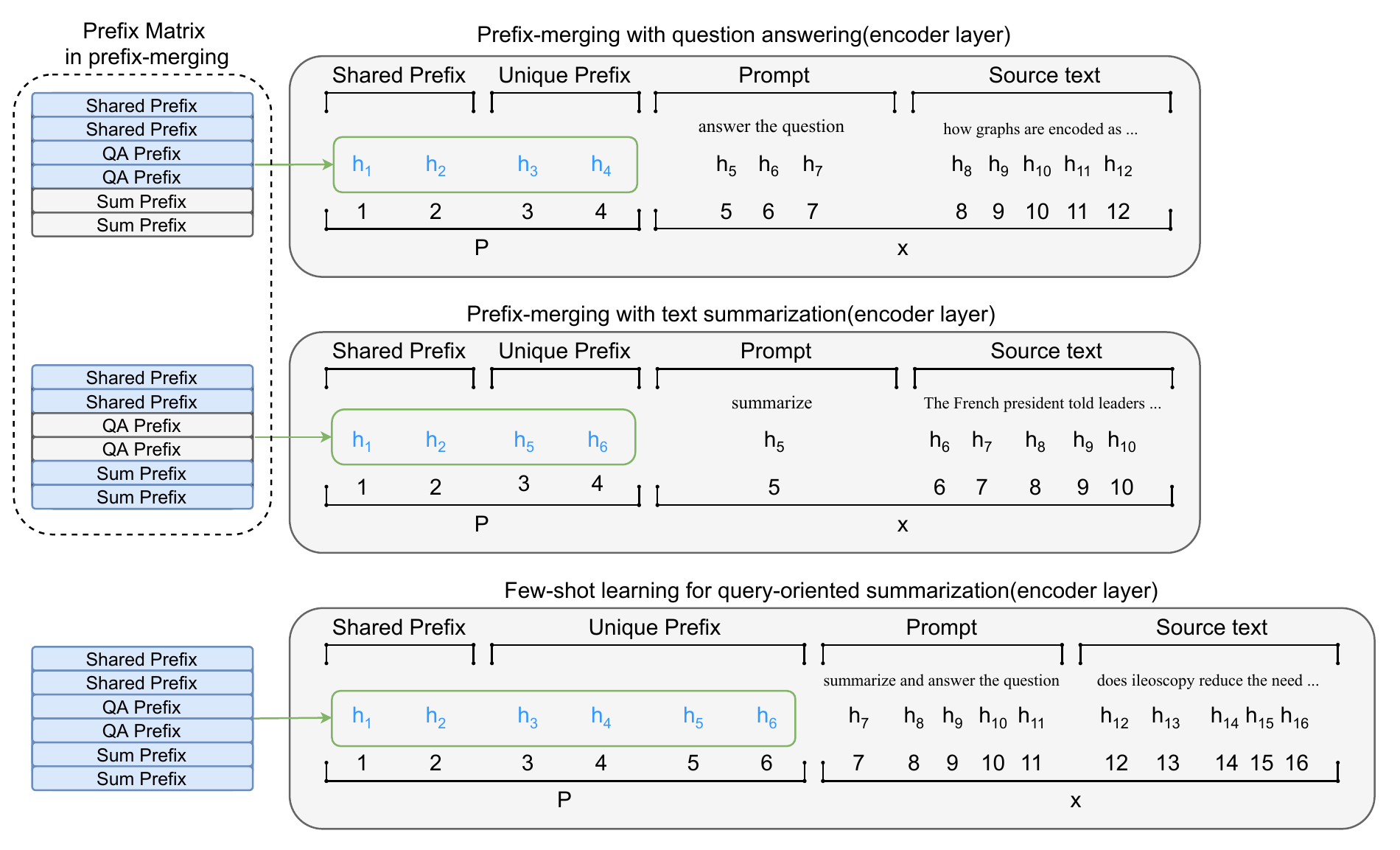}
	\caption{Focusing on the encoder layer of BART, the figure shows annotated examples and comparison between the prefix-merging (top, mid) on the two auxiliary tasks and applying the merged prefix on the target task with prefix-tuning (bottom).}
	\label{fig:label1}
\end{figure*}

\subsection{Problem Statement}

In this work, we aim to transfer the task-specific knowledge from text summarization and question answering (auxiliary tasks) to query-focused summarization (target task) to assist its learning. In this case, the query-focused summarization model can obtain a fair performance even with limited data. There are mainly two stages to accomplish this. In the first stage, a model is trained on the large-scale data from two auxiliary tasks to obtain the potentially useful knowledge for query-focused summarization. Here, we propose prefix-merging that merges task knowledge from auxiliary tasks into a particularly designed prefix. In the second stage, we train the model with data from query-focused summarization but with the assistance of the trained parameters from the first stage. For prefix-merging, the merged prefix is used to transfer the knowledge from the first stage to the second stage.

Our prefix-merging is considered as an extension of prefix-tuning, so we have a brief introduction about it in the section 3.2 as the background of our method. Then, we introduce our own method from section 3.3 to 3.5, and how we apply the merged prefix on query-focused summarization in 3.6.

\subsection{Prefix-tuning}
Consider there is a transformer-based encoder-decoder LM $p(y|x)$ such as Bart\cite{lewis2019bart} and it is parametrized by $\phi$. Taking the encoder layer in transformer as an example, let $z = [x]$ denote its input sequence. We use $h_i$ to represent the concatenation of all activation from all layers at the index $i$, and each activation consists of a key-value pair. The $h_i$ for all $i \in x $ in encoder layer is a function of $z_i$ and the other activations in the context based on the LM, as follows: 

\begin{equation}
h_i = LM_{\phi}(z_i,h_{\neq i})
\end{equation}

Prefix-tuning prepends a prefix for the encoder layer to obtain $z = [prefix;x]$, or prepends prefixes for cross-attention layer or self-attention layer in the decoder to obtain $z = [prefix;x;y]$ or $z = [prefix;y]$. Here, we use $P_{idx}$ to represent the sequence of prefix embedding indices, and $|P_{idx}|$ is used to represent the length of the prefix. A trainable matrix $P_{\theta} \in |P_{idx}| \times dim(h_i)$ is initialized to store the prefix parameters. Following the recurrence relation in equation (1), $h_i$ is calculated as below in prefix-tuning.

\begin{equation}
h_{i}=\left\{
\begin{array}{ll}
P_{\theta}[i,:], & if \; i \in P_{idx} \\
LM_{\phi}(z_i,h_{\neq i}), & otherwise
\end{array}
\right.
\end{equation}

Hence, $h_i$ becomes a function of the trainable $P_{\theta}$ and it allows the prefix parameters to control the model by affecting the activations in every layer of the transformer. 
During the training in prefix-tuning, the objective maintains the same as normal task, but only the prefix parameters $\theta$ are trainable and the parameters of the LM $\phi$ are fixed. In this case, the prefix parameters contain all the task-specific knowledge learned from the training.

\subsection{Intuition for Prefix-merging}

Intuitively, to merge the knowledge from different tasks into the prefix, the simplest way is to concatenate the individual prefix from these tasks. Another way is to use a shared prefix that is updated by all the tasks. Instead of using either of the two ways, we choose a more flexible prefix design for further investigation of the problem. For each task, its prefix consists of a shared sub-prefix (prefix embeddings shared by all tasks) and a task-specific sub-prefix (prefix embeddings used for a specific task) whose lengths are controlled by two hyperparameters. We believe the shared sub-prefix tends to represent the similarities between all merged tasks, while the task-specific sub-prefix refers to the uniqueness of each task. Meanwhile, the two mentioned intuitive methods can also be restored when any of the two hyperparameters is set to 0.

\subsection{Prefix-merging}
Similar to prefix-tuning, a trainable matrix $P_{\theta}$ is used to store the prefix parameters. The difference is that there are n different tasks denoted as $[task_1, task_2,..,task_n]$ that share or partly share the whole matrix. For each single task, it corresponds to several prefix embeddings in the prefix metrix, and we separate them into task-specific unique sub-prefix with a length of $l_{u}$ and a shared sub-prefix with a length of $l_{s}$. Figure \ref{fig:label1} shows an example of training two auxiliary tasks, text summarization and question answering, for prefix-merging. Here, both the shared sub-prefix length and unique sub-prefix length are set to 2. The prefix embedding indices for text summarization is [1,2,3,4], and it changes to [1,2,5,6] for QA.

In this way, the $P_{\theta}$ has the dimension of $(l_{s}+l_{u}*n) \times dim(h_i)$. We use $P_{idx}^n$ to represent the the sequence of prefix embedding indices of $task_n$ and its length $|P_{idx}^n|$ is equal to $l_{s}+l_{u}$. As follow, the $h_i$ for $task_n$ is calculated based on the following equation:

\begin{equation}
h_{i}=\left\{
\begin{array}{ll}
P_{\theta}[P_{idx}^n[i],:] , & if \; i  \leq |P_{idx}^n|\\
LM_{\phi}(z_i,h_{\neq i}) , & otherwise
\end{array}
\right.
\end{equation}

To distinguish the different tasks during the training, we add a task-specific prompt before the original input tokens following T5 \cite{raffel2019exploring}. As shown in Figure \ref{fig:label1}, the prompt is “summarize” for the text summarization and the prompt is “answer the question” for question answering. During the training, we adopt a mixed-task training strategy where instances from different tasks equally exist in the same training batch.

\subsection{Self-adaptive Prefix-merging}

Considering that manual design does not always lead to the best results, we further propose a self-adaptive prefix-merging. Instead of presetting the lengths of shared sub-prefix and unique sub-prefix, we aim to let the auxiliary tasks decide the prefix design. The idea is based on Fisher Information, a evaluation metric that reflects how much the model output changes when its parameters change. It can be considered as the importance of a parameter for the model on a certain set of data \cite{xu2021raise}. In this way, we can find the most important sub-prefix for each auxiliary task based on Fisher Information with the following equation:
\begin{equation}
    F_{i}=\frac{1}{pq} \sum_{j=1}^{p} \sum_{k=1}^{q}(\frac{\partial log(p(y_k|x_k;\theta))}{\partial \theta_j})^2
\end{equation}
where $F_(i)$ refers to the average Fisher information of the $i$-th prefix embedding, $p$ denotes the number of parameters in the embedding and $q$ represents the number of data. $x$ and $y$ refer to the input and output data in one auxiliary task.

During the training, we first initialize the prefix as a shared prefix trained by all auxiliary task for one epoch. Taking $task_n$ as an example, we then conduct a complete forward propagation and back propagation (one epoch) for all data in $task_n$, and calculate the Fisher Information for each prefix embedding. Only the top-$n$ prefix embeddings will be used in the later training for $task_n$ and others will be masked. In other words, the $P_{idx}^n$ is the indices of the top-$n$ prefix embeddings. After obtaining the important sub-prefix for each task, naturally, some prefix embeddings are shared by different tasks while others are task-specific. At last, we continue the training of the prefix on the auxiliary tasks with the selected sub-prefix.

\subsection{Applying the Merged Prefix to the Target Task}
After training on the auxiliary tasks, we obtain the prefix parameters that contain task knowledge from text summarization and question answering. We apply the knowledge to the target task, query-focused summarization, by using the merged prefix as initialization and continue prefix-tuning on it, but with a few differences. As shown in Figure  \ref{fig:label1}, all the prefix parameters are used for the target task including the shared sub-prefix and all the unique sub-prefixes. For self-adaptive prefix-merging, only the prefix embedding that is used for at least one auxiliary task is applied for the target task, otherwise it will be masked. We also adopt a new prompt that suggests the relation between the target task and auxiliary tasks. More specifically, we concatenate the prompt of text summarization and question answering as  “summarize and answer the question” for query-focused summarization.

\begin{table*}
	\centering
	\begin{tabular}{l|ccc|ccc|ccc}
		\hline \textbf{Data Size} &   & 50 &   &   & 150 &   &  & 300 &   \\
		\hline
		\hline
        \textbf{Model} & R-1 & R-2 & R-L & R-1 & R-2 & R-L & R-1 & R-2 & R-L \\
		\hline
		Random & 30.33 &  9.96 & 28.00   & 32.08 & 11.67 & 28.97   & 32.79 & 11.92 & 29.51\\
		Unq(30)        & 30.81 & 10.97 & 26.52                & 32.13 & 11.73 & 28.23            &  32.37 & 11.86 & 27.81\\
		Unq(20)+Sha(10)  & 32.36 & 11.40 & 28.30              & 33.14 & 12.12 & 29.10            & 33.68 & 12.39 & 29.81\\
		Unq(10)+Sha(20)   & 32.64 & 11.84 & \textbf{28.60}    & 33.46 & 12.34 & \textbf{29.46}   & 33.90 & 12.59 & \textbf{30.12}\\
		Sha(30)       & 32.44 & 11.48 & 28.17                 & 33.28 & 12.04 & 29.11            & 33.87 & 12.41 & 29.83\\	
		Self-adaptive & \textbf{33.18} & \textbf{12.01} & 28.45   & \textbf{33.66}  & \textbf{12.40}  &  28.98  &  \textbf{34.19} &  \textbf{12.65} & 29.53 \\	
		\hline
		BART(tar)  & 30.95 & 10.54 & 26.87              & 32.28 & 11.46 & 28.23            & 32.52 & 11.63 & 28.33\\		
		BART(aux+tar)  &  31.65  &  10.75  &  28.18    & 32.23 & 11.27 & 28.57     & 32.66 & 11.62 & 29.16\\
		BART\_base(full)  &  37.49  &  14.11  &  34.45    &  37.49  &  14.11  &  34.45   &  37.49  &  14.11  &  34.45 \\
		\hline
	\end{tabular}
	\caption{\label{font-table1}Evaluation result for query-focused summarization on PubMedQA. We compare the result on three different training data size: 50, 150, 300. Here, we also provide result of BART-base on the full-size training for better comparison.}
\end{table*}

\begin{table}
	\centering
	\begin{tabular}{l|ccc}
		\hline Model &  R-1  &  R-2  &  R-L \\ 
		\hline \hline 
		Random & 33.96 &  6.37 & 23.46 \\
		Unq(30) & 34.56  &  6.80  &  24.40 \\
		Unq(20)+Sha(10) & 34.54  &  6.64  &  23.53\\
		Unq(10)+Sha(20) &  34.87  &  7.23  &  24.70 \\
		Sha(30) & 34.53 & 6.89 & 23.98 \\
		Self-adaptive & \textbf{34.99} & \textbf{7.47} & \textbf{24.74}\\
		\hline
		BART(tar) & 33.83 & 6.52 & 22.71\\
		BART(aux+tar) & 12.85 & 4.42 & 15.17 \\
		Ext\_Oracle & 35.62 & 9.20 & 24.00 \\
		\hline
	\end{tabular}
	\caption{\label{font-table2}Evaluation result for query-focused summarization on DUC. Ext\_Oracle refers to oracle extractive result taking the query-related sentences as input, which can be seen as the upper bound of this experiment.}
\end{table}

\section{Experiment}
\subsection{Datasets}
To evaluate the idea of prefix-merging, we take query-focused summary as the target task, text summarization and question answering as two auxiliary tasks. We focus on commonly used datasets for query-focused summarization: PubMedQA and DUC. We also test our model on Debatepedia \cite{nema2017diversity} and have a discussion about it in the appendix. In terms of the PubMedQA, it requires the model to generate a summary containing 1-3 sentences as an answer to a question based on a medical related document. Since we train the target task under a few-shot situation, only part of the training set is used in the experiment and we test the model on the full testing set containing more than 20000 data samples. In terms of the DUC, it is a multi-document query-focused summarization dataset with hundreds of data samples. Hence, we adopt a extract-generate framework to conduct the experiment. We first adopt BM25 to extract a set of query-related sentences from the source documents and use the concatenation of the query and extracted sentences as the input of our model. The DUC 2006 is used for training and DUC 2007 is used for testing. In terms of the two auxiliary tasks, we adopt the XSum dataset \cite{narayan2018don}, a highly abstractive single-document summarization dataset, for the text summarization, and we use the classic machine reading comprehension dataset SQUAD 1.1 \cite{rajpurkar2016squad} for question answering.

\subsection{Experiment Setting}

Our implementation is based on the BART-large model from HuggingFace and all the input is truncated to 800 tokens. For the prefix-tuning based method, a default setting is a learning rate of $5 \times 10^{-5}$ and a prefix length of 30. The batch size is set to 48 when conducting prefix-merging, and for few-shot prefix-tuning, it changes with the size of the training data. In the experiment, we also use fine-tune based method as a comparison, and the default setting for it is a learning rate of $2 \times 10^{-5}$ and a batch size of 48. At training time, we adopt the AdamW optimizer with default hyperparameters. At inference time, we use beam search with a beam size of 2. The output length limitation is set from 30 to 75 tokens for PubMedQA and 250 to 300 for DUC. Since few-shot learning is sensitive to the training data, we train the models with three sets of training data and report the average result on PubMedQA. 

As for evaluation metric, following previous works, we apply ROUGE \cite{lin2004rouge} including Rouge-1 (R-1), Rouge-2 (R-2) and Rouge-L (R-L) for the query-focused summarization. We adopt a full Python implementation of the ROUGE-1.5.5, to conduct the experiment.

\subsection{Result}

We first evaluate the different prefix designs within three different few-shot learning data sizes (50, 150, 300) for the target task in Table \ref{font-table1}. "Unq(n)" stands for the total number of the prefix embeddings in all unique sub-prefix, while "Sha(n)" refer to the shared sub-prefix. For example, "Unq(10)+Sha(20)" represent the merged prefix consists of unique sub-prefix with length 10 (5 for each task) and the shared sub-prefix with length 20. In terms of the self-adaptive prefix-merging, we initialize the prefix length as 40 and select the top-25 prefix embeddings for each tasks. In this case, self-adaptive prefix-merging is more likely to have a comparable parameter numbers with the other prefix designs, which makes a fair comparison. We also add a baseline “random”: randomly initialize the prefix and conduct few-shot prefix-tuning on the query-focused summarization dataset. We further compare our model with BART in three training settings:(1) BART(tar) refers to fine-tuning the BART only use the limited data from query-focused summarization; (2) BART(aux+tar) refers to first fine-tuning on the auxiliary tasks then fine-tuning on query-focused summarization, which is similar to some previous approaches \cite{yu2021adaptsum} and \cite{fabbri2020improving}; (3) BART(full) refers to fine-tuning on the large-scale data from query-focused summarization.

\begin{table}
	\centering
	\begin{tabular}{l|ccc}
		\hline Model &  R-1  &  R-2  &  R-L \\ 
		\hline \hline
		Fine+Fine & 31.65  &  10.75  &  28.18 \\
		Fine+Prefix & 31.64  &  10.79  &  27.57\\
		Prefix+Fine &  32.03  &  11.30  &  28.12 \\
		Prefix+Prefix & \textbf{33.18} & \textbf{12.01} & \textbf{28.45}\\
		\hline
	\end{tabular}
	\caption{\label{font-table2}The comparison between prefix-merging and fine-tuning with a training data size of 50.}
\end{table}

In Table \ref{font-table2}, we compare the prefix-merging with fine-tuning. Since it is a two-stage training process (training on auxiliary tasks then applying on the target task), each stage can adopt prefix-based training (only the prefix parameters are trained and the LM parameters are frozen) or fine-tuning (all parameters are trained). Therefore, we report four variants in total: (1) fine-tuning + fine-tuning (Fine+Fine), which is the same as BART(aux+tar); (2) fine-tuning + prefix-tuning (Fine+Prefix); (3) prefix-merging + fine-tuning (Prefix+Fine); (4) prefix-merging + prefix-tuning (Prefix+Prefix), which is our proposed approach in Section 3. Despite the variant (1), we add a prefix of length 30 to the model. Taking variant (2) as an example, firstly, both the prefix and the LM are updated by the training data from auxiliary tasks and then only the prefix parameter is trained on the target task.

Table \ref{font-table3} displays the result of using different auxiliary tasks for query-focused summarization. “Sum+QA” refers to the best result when using both text summarization and QA; “Only Sum” and “only QA” are designed for ablation study where only one of the two tasks is used in stage one. Moreover, we also import a baseline “Unrelated Task” that takes sentence copying as the auxiliary task, which contains no useful task knowledge for query-focused summarization. We use prefix-tuning to train the model when there is only one auxiliary task.

We summarize the experiment result with the following conclusions.

\begin{table}
	\centering
	\begin{tabular}{l|ccc}
		\hline Model &  R-1  &  R-2  &  R-L \\ 
		\hline \hline
		Unrelated Task & 31.34  &  10.77  &  27.08 \\
		Only Sum  &  32.38  &  11.56  &  27.75 \\
		Only QA & 31.78  &  11.39  &  28.43 \\
		Sum and QA & \textbf{33.18} & \textbf{12.01} & \textbf{28.45}\\
		\hline
	\end{tabular}
	\caption{\label{font-table3}The comparison between using different auxiliary tasks with a training data size of 50.}
\end{table}

\begin{figure*}[ht]
	\centering
	\includegraphics[scale=0.51]{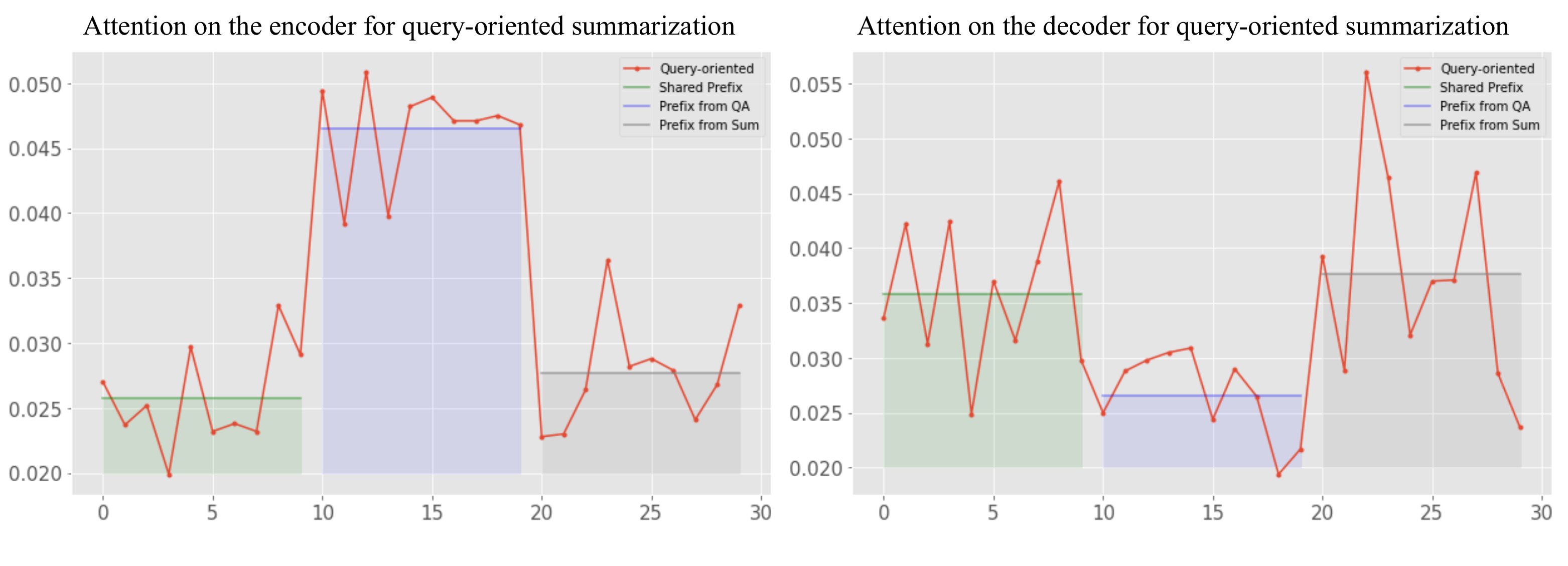}
	\caption{The attention score for query-focused summarization in both encoder and decoder of model “Unq(20)+Sha(10)”.}
	\label{fig:label2}
\end{figure*}

\textbf{The self-adaptive prefix-merging achieves a comparable result with the best manually prefix design.} It is not a surprise that self-adaptive prefix-merging outperforms most of the prefix designs and achieves the best result in both datasets. One thing that is worth noticing is that the effective length for self-adaptive prefix-merging is also around 30 (initialized as 40 and 10 are masked by all tasks), which means the number of parameter maintains equal with other prefix design. Meanwhile, its proportion of shared sub-prefix and unique sub-prefix is similar to the best manual design Unq(10)+Sha(20). This suggests that self-adaptive prefix-merging has the ability to find the best prefix design automatically. Compared with BART, self-adaptive prefix-merging outperforms both BART(tar) and BART(aux+tar), which indicates the effectiveness of prefix-merging. In the experiment on DUC, we notice that BART(aux+tar) drops a lot compared with other results. We believe this is because the difference between DUC and datasets used in auxiliary tasks is relatively huge and the generalization ability of BART is lost after training on the auxiliary tasks.

\textbf{Prefix-merging is better than fine-tuning for integrating and transferring task knowledge to the downstream task.}
In Table 2, prefix-merging outperforms fine-tuning with both downstream training approaches. On the one hand, this is because the generalization ability of the LM is preserved when its parameters are frozen. On the other hand, we believe using prefix as the container of new task knowledge is more similar to the natural form of LM. We believe this shows the potential of prefix-merging in many-to-one knowledge transferring.

\textbf{The merged prefix contains effective task knowledge from both auxiliary tasks.}
The initialization of prefix is believed to have a huge effect on the prefix-tuning based approaches. Here, “unrelated task” stands for the performance when the prefix is well-initialized while containing no knowledge for the target task. Compared to it, using one auxiliary task, either text summarization or QA, achieve a better result. This suggests that the two tasks contribute useful knowledge to query-focused summarization. More importantly, prefix-merging gets the best performance. And this can be achieved only when the prefix-merging allows the prefix to integrate effective task knowledge from both tasks.

\begin{table}
	\centering
	\begin{tabular}{l|ccc}
		\hline Model &  R-1  &  R-2  &  R-L \\ 
		\hline \hline
		--Prefix  & 26.56  &  8.19  &  22.16\\
		--Prompt  & 32.48  &  11.63  & 28.57 \\
		Unq(10)+Sha(20) & \textbf{32.64} & \textbf{11.84} & \textbf{28.60}\\
		\hline
		Sha(40)  & 32.60  &  11.74  & \textbf{28.54} \\
		Self-adaptive & \textbf{33.18} & \textbf{12.01} & 28.45\\
		\hline
	\end{tabular}
	\caption{\label{font-table4}The experiment result for ablation study with a training data size of 50.}
\end{table}

\subsection{Ablation Study}
For more detailed analysis, we design an experiment to explore how different components contribute to our approach. We remove the prefix (-prefix) and the prompt (-prompt) from during the training of the query-focused summarization. The prefix design used here is Unq(10)+Sha(20). We can observe that removing the prompt has a small negative influence on the result. We believe this is because the input form of text summarization and QA is different and the model can distinguish the two tasks even without the given prompt. We also find that the performance drops a lot once the prefix is removed. This indicates that the prompt only plays as guidance, while the prefix is the one containing the task-specific knowledge. For self-adaptive prefix-merging, we compare it with its base prefix design without self-adaption, Sha(40). Even with more trainable parameters, self-adaptive prefix-merging still outperforms it. The result shows that prefix embeddings selected by Fisher Information are crucial for the tasks.

\subsection{Prefix Visualization}
To have a more direct observation, we visualize the attention on the prefix during the inference for query-focused summarization in Figure \ref{fig:label2}. We adopt the attention weights passing through the Softmax layer and further normalize the attention weights only on the prefix embeddings. The final attention score is obtained by averaging attentions from all heads in all layers from 100 random samples. In Figure \ref{fig:label2}, the x axis refers to the indices of the prefix embedding and y axis is the normalized attention score. The straight lines with colors stand for the position of the three types of sub-prefix, shared sub-prefix (0-9), unique sub-prefix originated from QA (10-19) and unique sub-prefix originated from Summarization (20-29), and their heights refer to the average attention score, which can be considered as the prefix's contribution to the query-focused summarization. In this case, it explains how the merged prefix works for query-focused summarization. 

For the decoder, we display the attention in the cross-attention layer. In terms of the encoder, since the model needs to understand the query, we believe it is reasonable that the sub-prefix originated from QA plays the most important role. In terms of the decoder, the sub-prefix originated from QA has little effect on the model, while the shared sub-prefix and sub-prefix originated from summarization dominate. This is because generating the query-focused summaries relies more on generation ability and summarization ability. These findings suggest that the knowledge from QA and summarization is properly used for query-focused summarization through the merged prefix.

\section{Conclusion}
In this paper, we show that prefix-merging is an effective approach for transferring and integrating task knowledge from multiple auxiliary tasks to a target task with limited data. In the context of query-focused summarization, integrating text summarization and QA, our approach outperforms the traditional approach fine-tuning. We further discuss the influence of different prefix designs and propose a self-adaptive prefix-merging. We also provide a visualize explanation for how the merged prefix works. Although this paper focuses on query-focused summarization, we believe these findings suggest a new application for prompt-based approaches in multi-task situation, providing guidance for future progress in this field.

\section{Limitations}
Prefix-merging is based on a seq2seq pretrained model, Bart, so it is hard for our model to deal with long input that exceed the input limitation of the pretrained model. Hence, we mainly focus on single-document query-focused summarization. In terms of the experiment, unfortunately, there is seldom few-shot single-document query-focused summarization model. Although there exist some multi-document query-focused summarization models with weak supervision\cite{xu2020generating,laskar2020wsl,xu2020coarse}, these models all follow a coarse-to-fine framework, which make them hard to directly compared with our model. Hence, we mainly use BART with different training settings as comparison and focus more on the longitudinal comparison. Moreover, we believe that prefix-merging has the potential to be used for other complex tasks that can be integrated from basic tasks. However, we only finish the research in the context of query-focused summarization, which leaves future direction for our work. 

\section*{Acknowledgements}
The work described in this paper was supported by Research Grants Council of Hong Kong (PolyU/15203617 and PolyU/5210919), National Natural Science Foundation of China (61672445, 62106165).

\bibliography{anthology,custom}
\bibliographystyle{acl_natbib}

\appendix
\section{Appendix}
\subsection{Experiment on Debatepedia}
\label{sec:appendix}
Debatepedia \cite{nema2017diversity} is one of the commonly used query-focused summarization dataset. However, during our experiment, we find a serious but unintentional data leakage problem between the training set and the testing set in its standard division. Around 64\% of summaries in the testing set appear or have similar ones in the training set (difference is lower than 2 words). In this case, the model tends to remember the data samples rather learning to do query-focused summarization.

\begin{table*}
	\centering
	\begin{tabular}{l|ccc|ccc|ccc}
		\hline \textbf{Data Size} &   & 50 &   &   & 150 &   &  & 300 &   \\
		\hline
		\hline
        \textbf{Model} & R-1 & R-2 & R-L & R-1 & R-2 & R-L & R-1 & R-2 & R-L \\
		\hline
		Random & 0.20 & 0.30 & 0.45 & 0.30 & 0.22 & 0.36 & 0.08 & 0.11 & 0.45\\
		Unq(30)        & 0.45 & 0.23 & 0.40 & 0.41 & 0.36 & 1.02 & 0.08 & 0.08 & 0.11\\
		Unq(20)+Sha(10)  & 0.50 & 0.22 & 0.25 & 0.04 & 0.22 & 0.59 & 0.17 & 0.09 & 0.23\\
		Unq(10)+Sha(20)   & 0.14 & 0.01 & 0.13 & 0.23 & 0.15 & 0.16 & 0.15 & 0.02 & 0.32\\
		Sha(30)       & 0.19 & 0.08 & 0.18 & 0.25 & 0.27 & 0.64 & 0.23 & 0.10 & 0.24\\	
		Self-adaptive & 0.38 & 0.16 & 0.48 & 0.10 & 0.20 & 0.71 & 0.12 & 0.08 & 0.35 \\	
		\hline
		BART(tar)  & 0.53 & 0.52 & 0.51 & 0.21 & 0.15 & 0.25 & 0.32 & 0.22 & 0.14\\		
		BART(aux+tar)  &  0.47 & 0.32 & 0.43 & 0.24 & 0.17 & 0.21 & 0.18 & 0.14 & 0.16\\
		\hline
	\end{tabular}
	\caption{\label{font-table7}.Standard Deviation of the Results on PubMedQA}
\end{table*}

\begin{table}[h]
	\centering
	\begin{tabular}{l|ccc}
		\hline Model &  R-1  &  R-2  &  R-L \\ 
		\hline \hline 
		Random & 18.57 & 5.50 & 17.50 \\
		Unq(30) & 21.53  &  6.77  &  20.07 \\
		Unq(20)+Sha(10) & 22.20  &  7.20  &  \textbf{20.59}\\
		Unq(10)+Sha(20) &  22.03  &  7.13  &  20.29 \\
		Sha(30) & 21.85 & 6.92 & 20.27 \\
		Self-adaptive & \textbf{22.29} & \textbf{7.39} & 20.53\\
		\hline
		BART(tar) & 21.60 & 6.81 & 19.89\\
		BART(aux+tar) & 21.36 & 6.24 & 19.00 \\
		\hline
		BART(full) & 57.74 & 43.42 & 57.03\\
		BART(full)\_redivided & 24.80 & 8.06 & 22.95 \\
		\hline
	\end{tabular}
	\caption{\label{font-table6}Evaluation result on Debatepedia.}
\end{table}

This observation is also supported by the experiment. In the upper and middle part of Table \ref{font-table6}, we display the result of few-shot learning with 50 data samples on standard division of Debatepedia. In the lower part of Table \ref{font-table6}, we show the result of BART training with full-size data but with different data division. BART(full) represents the standard division and BART(full)\_redivided refers to a new division that do not have the data leakage problem (we achieve this by redivide all data samples by an alphabetical sort, where similar data samples tend to gather together rather than scatter in both training and testing set). We observe a huge gap between the BART(full) and BART(full)\_redivided and it can not be explained by the difference of the division. Meanwhile, the result of few-shot learning is much lower than the result of full-size training. Both phenomenon suggest there exist a data leakage problem. The poor performance of BART on the redivided Debatepedia also make us question whether Debatepedia is qualified for query-focused summarization. Hence, we discuss this problem in the appendix and hope more researchers can notice this. 

\subsection{Standard Deviation of the Results on PubMedQA}
To have a better understanding of the experiment results on PubMedQA, we report the standard deviation (std) across multiple runs in the experiment on PubMedQA in Table \ref{font-table7}.

\end{document}